%% file: main.tex
\documentclass{article}

\usepackage[final, nonatbib]{neurips_2024_ml4ps}

\usepackage[utf8]{inputenc} 
\usepackage[T1]{fontenc}    
\usepackage{hyperref}       
\usepackage{url}            
\usepackage{booktabs}       
\usepackage{amsfonts}       
\usepackage{nicefrac}       
\usepackage{microtype}      
\input{pakete}
\input{befehle}

\input{makros}

\usepackage[linesnumbered,ruled]{algorithm2e}

\usepackage{algorithmicx}
\title{Explicit and data-Efficient Encoding via Gradient Flow}

\author{Kyriakos Flouris \\ 
Computer Vision Laboratory, ETH Zurich \\
\texttt{kflouris@vision.ee.ethz.ch}
\And
Anna Volokitin \\
Computer Vision Laboratory, ETH Zurich \\
\texttt{voanna@vision.ee.ethz.ch}
\AND
Gustav Bredell \\
Computer Vision Laboratory, ETH Zurich \\
\texttt{gustav.bredell@vision.ee.ethz.ch}
\AND
Ender Konukoglu \\
Computer Vision Laboratory, ETH Zurich \\
\texttt{ender.konukoglu@vision.ee.ethz.ch}
\AND
}

\begin{document}

\maketitle

\begin{abstract}
The autoencoder model typically uses an encoder to map data to a lower dimensional latent space and a decoder to reconstruct it. However, relying on an encoder for inversion can lead to suboptimal representations, particularly limiting in physical sciences where precision is key. We introduce a decoder-only method using gradient flow to directly encode data into the latent space, defined by ordinary differential equations (ODEs). This approach eliminates the need for approximate encoder inversion. We train the decoder via the adjoint method and show that costly integrals can be avoided with minimal accuracy loss. Additionally, we propose a $2^{nd}$ order ODE variant, approximating Nesterov's accelerated gradient descent for faster convergence. To handle stiff ODEs, we use an adaptive solver that prioritizes loss minimization, improving robustness. Compared to traditional autoencoders, our method demonstrates explicit encoding and superior data efficiency, which is crucial for data-scarce scenarios in the physical sciences. Furthermore, this work paves the way for integrating machine learning into scientific workflows, where precise and efficient encoding is critical. \footnote{The code for this work is available at \url{https://github.com/k-flouris/gfe}.}

\end{abstract}

\section{Introduction}

Auto-encoders are widely successful in supervised learning due to their ability to learn lower-dimensional representations of input data, enabling efficient computation \cite{stackedAE}. The basic idea is that when sufficient correlation exists within the input data, the latent dimension can generate a model of the input. However, since the encoder is not directly learned, the encoding process can be semi-arbitrary, leading to suboptimal latent space representations and inefficient learning \cite{deepsdf}. To address this, efforts have been made to directly regularize the latent space \cite{Tschannen2018RecentAI}. Typically, the encoder approximates the inverse of the decoder, but this requires learning additional parameters, impeding learning, particularly with limited data. Flow models \cite{DBLP:journals/corr/DinhKB14,NEURIPS2023_572a6f16} attempt to resolve this by using invertible maps, but they are restricted to equi-dimensional latent spaces. What if we could eliminate the encoder, retain the advantages of a lower-dimensional latent space, and directly optimize the representation? This could enable faithful reconstruction with fewer samples and iterations.

Currently, DeepSDF \cite{deepsdf} optimizes latent space representations using a continuous signed distance function to model 3D shapes, where a latent vector $z$ represents the shape encoding. Hamiltonian Neural Networks \cite{greydanus2019hamiltonian} and VampPrior \cite{tomczak2018vae} introduce physically-informed dynamics and more expressive priors for VAEs, respectively, contributing to latent space optimization. SAVAE \cite{kim2018semi} and the energy-based model approach in \cite{pang2020learninglatentspaceenergybased} directly aim to optimize the encoder through iterative refinement.

In this work, we propose a novel encoding method, namely gradient flow encoding (GFE). This decoder-only method explicitly determines the optimal latent space representation at each training step via a gradient flow, namely, an ordinary differential equation (ODE) solver. The decoder is updated by minimizing the loss between the input image and its reconstruction from the optimal latent space representation. Although slower, this method converges faster and demonstrates superior data efficiency and ultimately achieves better reconstructions with a minimal number of training samples compared to a conventional auto-encoder (AE). A notable advantage of GFE is the reduction in network size, as it eliminates the need for an encoder.

Using a gradient flow for encoding can be computationally demanding. ODE solvers with adaptive step sizes aim to minimize integration error, which is crucial for accuracy. However, these solvers can slow down training when the ODE becomes stiff, as step sizes shrink and make integration time-consuming \cite{chen2018neural, grathwohl2019ffjord}. This is especially problematic in GFE-based training.
We find that the exact gradient flow path may be less important than reaching convergence, so minimizing integration error may not be optimal. Fixed step sizes also cause poor convergence and stability in decoder training. To address this, we develop an adaptive solver that adjusts each step to minimize the loss between input and output. We also introduce a loss convergence assertion in the adaptive minimize distance (AMD) solver, making gradient flow more computationally feasible for neural networks.

 For optimizing the latent space and decoder in GFE, we implement an adjoint method \cite{butcher2008numerical}, which has also been used in other recent works \cite{chen2018neural}. To improve efficiency, we demonstrate that a full adjoint method may not always be necessary, and an approximation can be effectively utilized. Moreover, we introduce a Nesterov second-order ODE solver, which accelerates convergence for each training data size. Ultimately, the approximate GFE utilizing AMD (GFE-amd) is employed for testing and comparison with a traditional AE solver.  Our flow-inspired model offers a solution for the stringent data efficiency and robustness demands in fields like physics, astronomy, and materials science. By utilizing the flow-model for encoding, we contribute to algorithms that create interpretable, accurate predictive models, enabling more robust discoveries and insights in scientific research.

\section{Method}

An auto-encoder maps an input $y$ to a lower-dimensional latent space $z$ via an encoder $E$, then reconstructs it using a decoder $D$, with $E$ and $D$ acting as approximate inverses. During training, each sample is mapped to $z$ by $E$, reconstructed by $D$, and the reconstruction error is minimized with respect to both networks' parameters. At any point, an optimal latent representation $z^*$ minimizes the reconstruction error, but the encoder doesn't directly map to $z^*$. Instead, $E(z)$ is updated to make its reconstruction closer to the sample, which may not be efficient. Alternatively, $z^*$ can be found and used to update the decoder directly.

 Determination of $z^*$ for each sample $y$ can be formulated as an optimization problem: $z^* = \arg\min_z l(y, D(z,\theta)),$
where $\theta$ represents the parameters of the decoder network and $l(\cdot,\cdot)$ is a distance function between the sample and its reconstruction by the decoder, which can be defined based on the application. One common form is the $L_2$ distance $\lVert D(z,\theta) - y \rVert^2_2$. The optimization can be achieved by gradient descent minimization. To integrate this minimization into the training of the decoder network, a continuous gradient descent algorithm is implemented via the solution to an ODE:
\begin{equation}
\frac{dz}{dt} = -\alpha(t) \nabla_z l(y, D(z(t),\theta)), \quad   z(0) = 0,
\label{eq:ode}
\end{equation}
where time $t$ is the continuous parameter describing the extent of the minimization, and $\alpha(t)$ is a scaling factor that can vary with time. When the extremum is reached, $z$ reaches a steady state. In practice, we compute the optimal $z^*$ by integrating the ODE:
\begin{equation}
z^* = z(\tau) = \int_0^{\tau} -\alpha(t)\nabla_z l(y, D(z(t), \theta)) \, dt, \quad z(0) = z_0,
\end{equation}
where $z_0$ is the initialization of the optimization, which is set as the zero vector in our experiments.
Consequently, after minimizing $z \rightarrow z^* \equiv z(t=\tau)$ for a given decoder $D$ (the 'forward model'), the decoder is trained with a total loss function for a given training set $M$ (the 'backward step'):
\begin{equation}
\mathcal{L}(\theta) = \sum_{m=1}^M l(y_m, D(z_m^*,\theta)).
\end{equation}
At each iteration, while searching for $\arg\min_{\theta} \mathcal{L}(\theta)$, a new $z^*$ is recalculated for each sample. $\arg\min_{\theta} \mathcal{L}(\theta)$ is computed via the standard adjoint method, as explained in App. \ref{app:adjoint}.

\subsection{Nesterov's 2nd-Order Accelerated Gradient Flow}
\label{sec:nesterov}

The gradient flow described above is based on naive gradient descent, which may be slow in convergence. The convergence per iteration can be further increased by considering Nesterov's accelerated gradient descent. A second-order differential equation approximating Nesterov's accelerated gradient method has been developed in \cite{nesterov2016}, and additionally incorporated into Adam \cite{dozat.2016}. This $2^{nd}$ order ODE for $z$ is given by:
\begin{equation}
    \frac{d^2 z}{dt^2} + \frac{3}{t} \frac{dz}{dt} + \nabla_z l(y, D(z, \theta)) = 0,
\end{equation}
with the initial conditions $dz/dt\small|_{t=0} = z(0) = 0$. 
To incorporate this within the gradient flow encoding framework, we split the $2^{nd}$ order ODE into two interacting $1^{st}$ order equations and solve them simultaneously. Specifically, we solve:
\begin{equation}
   \frac{dv}{dt} = -\frac{3}{t + \epsilon}v + \nabla_z l(y, D(z, \theta)), \quad \frac{dz}{dt} = v,
\end{equation}
where $\epsilon$ ensures stability at small $t$.

\subsection{Adaptive minimise distance solver}
\label{sec:amd}

Step size is crucial in  discretized optimization algorithms for reaching a local extremum, including solving the gradient flow equation for optimal latent space representation. Fixed time-step solvers can cause instabilities during decoder training due to stiffness, as predefined time slices may not adapt to rapid changes in the gradient flow, see App. \ref{app:fixed}. For example, a $4^{th}$ order Runge-Kutta method with a fixed grid uses $\d t$ slices in a logarithmic series to manage variations near $t=0$, but this can still lead to instability, particularly in forward and backward passes. While adaptive step-size ODE solvers can theoretically address these issues, stiffness in the gradient flow equation remains a challenge.

Adaptive step-size ODE solvers address stiffness in gradient flow equations but prioritize accurate integration, which isn't always useful for decoder training. A more effective approach minimizes loss at each step, regardless of the integration path. To achieve this, we develop an adaptive step-size method resembling an explicit ODE solver, like the forward Euler method, but without a fixed grid. The challenge is solving ODE \ref{eq:ode} while selecting time-steps that reduce $l(y, D(z(t), \theta))$. Essentially, this is gradient descent with adaptive step-size selection~\cite{bertsekas_2016}. In this framework, $\alpha(t)$ is redundant and can be set to 1, with the time-step $\delta t$ taking over its role.

In the AMD method, at each time $t_n$, $\delta t_n$ is chosen by finding the smallest $m = 0,1,...$ that satisfies:
\begin{equation}
     l\left(y, D(z(t_n) - \beta^m s_n \nabla_z l(y, D(z(t_n), \theta)), \theta)\right) < l\left(y, D(z(t_n), \theta)\right),
\end{equation}
with $\beta \in (0,1)$ (set to 0.75 in our experiments) and $s_n$ as a scaling factor. At each time point $t_n$, the time-step is set as $\delta t_n = \beta^m s_n$. The scaling factor is updated at each iteration as $\hat{s}_n = \max(\kappa s_{n-1}, s_0),\ s_n = \min(\hat{s}_n, s_{max}),\ s_{max}=10,\ s_0 = 1,\ \kappa = 1.1$. Based on this, $t_{n+1} = t_n + \delta t_n$ and
\begin{equation}
     z(t_{n+1}) = z(t_n) - \delta t_n \nabla_z l(y, D(z(t_n), \theta)).
\end{equation}
If the chosen time-step goes beyond $\tau$, a smaller step is used to reach $\tau$ exactly. The solution of the integral of Eq. \ref{eq:ode} is then $z(\tau)$. Furthermore, the AMD solver monitors the gradient of the convergence curve to determine if the loss function is sufficiently optimized, allowing it to assign a new final $\tau'$ and stop early to avoid unnecessary integration.

\section{Results and Discussion}

\begin{figure}[ht]
\begin{center}
\includegraphics[width=\columnwidth]{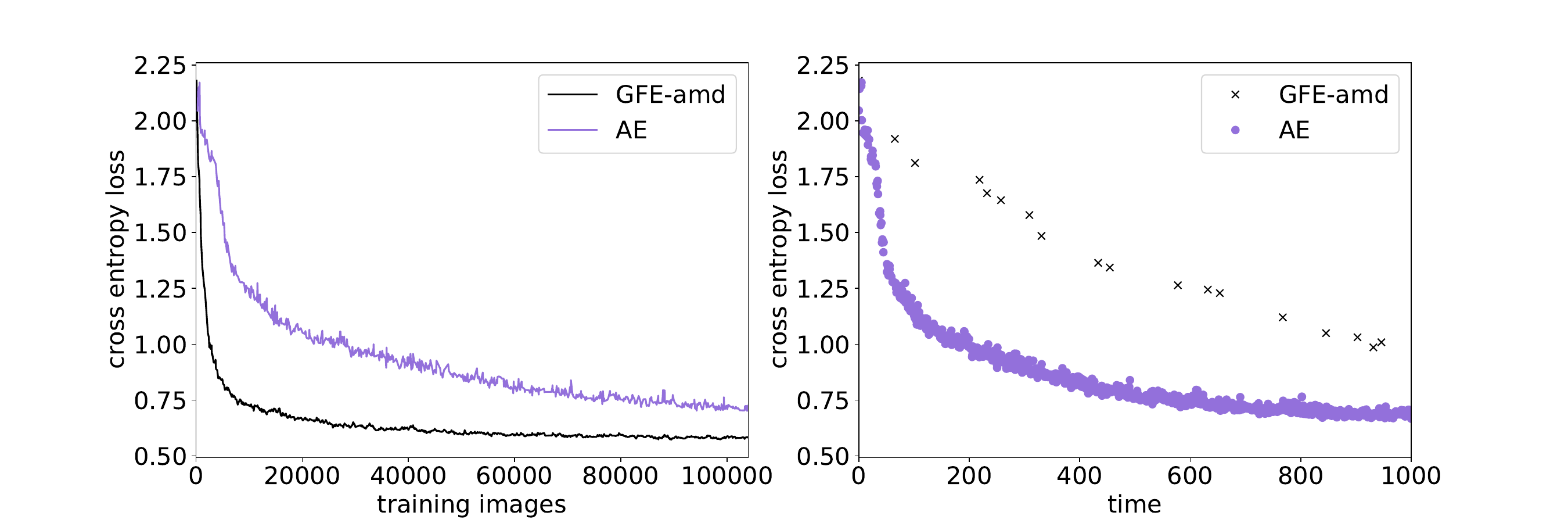}
\end{center}
\caption{\label{fig:GFE_AE_step_relative} \textbf{Left} Validation mean cross-entropy loss vs. number of MNIST training images for GFE-amd and AE methods, with GFE-amd showing significant convergence with minimal training data. \textbf{Right} Validation mean cross-entropy loss vs. time for GFE-amd and AE methods, with AE being faster due to more iterations in the same time span.}
\end{figure}

A direct comparison of GFE-amd to a conventional AE for MNIST training is shown in Fig. \ref{fig:GFE_AE_step_relative}. Further experimental details can be found in App. \ref{app:experiments}. The x-axis shows the number of training images processed, not iterations. The training uses mini-batch data with replacement, revisiting the data multiple times. GFE-amd demonstrates significantly better learning per image, nearly converging at 800,000 images, see Fig. \ref{fig:GFE_AE_step_relative}~left.  This is a consequence of efficient latent space optimization. However, this comes with higher computational cost per iteration due to the ODE solver, see Fig. \ref{fig:GFE_AE_step_relative}~right. Both models use the Adam optimizer, therefore the difference can be attributed to better gradients the GFE-amd model generates to update the decoder network at each training iteration.

This increase in computation is not necessarily a disadvantage, considering the efficient learning of the GFE-amd method. In Table \ref{tb:results} right, the average cross entropy loss for a complete test-set is recorded for both methods for some small number of training images. The GFE-amd is able to learn quite well even after seeing a tiny fraction of the total training data. Furthermore, the GFE-amd method noticeably improves an AE trained decoder when it is used to test, the result of an optimized latent space even without a network parameters update.



To verify the overall quality of the method both the AE and GFE-amd are tested when converged as shown in Table \ref{tb:results} left. 
The GFE-amd performs very similar to AE both for MNIST, Segmented MNIST (SegMNIST) and FMNIST. 
It is worth noting that the GFE-amd trainings are on average converged at $1/12^{th}$ of the number of iterations relative to the AE. 
For the  SegMNIST the networks are fully trained while seeing only the first half (0-4) of the MNIST labels and they are tested with the second half (5-9) of the labels. The GFE-amd shows a clear advantage over the AE emphasizing the versatility of a GFE-amd trained neural network.

The interpretable nature of the proposed method inherently supports scalability, suggesting that it will remain data-efficient even when applied to real-world or large-scale datasets.

\begin{table}[ht]
\centering
\caption{Average cross-entropy loss: \textbf{Left} Table shows different MNIST training sizes (percentage is of total) the mixed column refers to AE training and then GFE-amd testing. \textbf{Right} Table shows complete training tests results on different datasets. \label{tb:results}}
\begin{tabular}{c c c c} 
\toprule
Training Images & AE & GFE-amd & Mixed \\ [0.5ex] 
\midrule
480 (0.24\%)  & 0.2660 & \textbf{0.2098}  & 0.2634 \\ 
1920 (0.98\%) & 0.2488 & \textbf{0.1558}  & 0.2323 \\ 
5760 (2.9\%)  & 0.1954 & \textbf{0.1136}  & 0.1829 \\ 
\bottomrule
\end{tabular}
\hspace{0.5cm}
\begin{tabular}{c c c} 
\toprule
Dataset & AE & GFE-amd \\ [0.5ex] 
\midrule
MNIST         & 0.0843 & \textbf{0.0830} \\ 
SegMNIST & 0.1205 & \textbf{0.1135} \\ 
FMNIST        & \textbf{0.2752} & \textbf{0.2764} \\ 
\bottomrule
\end{tabular}

\end{table}

\section{Conclusions}
To this end, a gradient flow encoding, decoder-only method was investigated. The decoder-dependent gradient flow searches for the optimal latent space representation, which eliminates the need for an approximate inversion. The full adjoint solution and its approximation are leveraged for training and compared in various settings. Furthermore, we present a $2^{nd}$ order ODE variant to the method, which approximates Nesterov's accelerated gradient descent, with faster convergence per iteration. Additionally, an adaptive solver that prioritizes minimizing loss at each integration step is described and utilized for comparative tests against the autoencoding model. While each training step takes longer, the gradient flow encoding converges faster and demonstrates much higher data efficiency than the autoencoding model. This flow-inspired approach opens up new possibilities for efficient and robust data representations in scientific fields such as physics and materials science, where accurate modeling and data efficiency are paramount.

\section*{Acknowledgement}

This project was supported by grants \#2022-531 and  \#2022-643 of the Strategic Focus Area "Personalized
Health and Related Technologies (PHRT)" of the ETH Domain (Swiss Federal Institutes of
Technology).

\bibliographystyle{abbrv}
\bibliography{main}

\appendix

\section{Method details}
\subsection{The adjoint method for the gradient flow}
\label{app:adjoint}
 As described above, after finding,  $z^* \equiv z(\t) =\arg_z\min l(y, D(z,\th))$ the total loss is minimized with respect to the model parameters. 
 The dependence of $z^*$ to $\th$ creates an additional dependence of $\mathcal{L}(\th)$ to $\th$ via $z^*$.
 For simplicity, let us consider the cost of only one sample $y$, effectively $l(y, D(z^*, \th))$. We will compute the total derivative $d_\th l(y, D(z^*, \th))$ for this sample, and the derivative of the total cost for a batch of samples can be computed as the sum of the sample derivatives in the batch.
 The total derivative $d_\th l(y, D(z^*, \th))$ is computed as
\begin{equation*}
    d_\theta l(y, D(z^*, \th)) = \partial_\theta l(y, D(z^*, \th)) + \partial_{z^*}l(y, D(z^*, \th))\partial_\theta z^*.
\end{equation*}
The derivative $\partial_{z^*}l(y, D(z^*, \th))d_\theta z^*$ can be computed using the adjoint method and leads to the following set of equations
\begin{align}
\label{eq:adjoint1}
d z/d t = - \a(t) \nabla_z l(y, D(z(t), \th)), \text{ with } z(0)=0 \\
\label{eq:adjoint2}
d\l/d t  = - \a(t) \l^T  \nabla^2_z l(y, D(z(t), \th)), \text{ with } \l(\tau) = -\nabla_z l(y, D(z(\t), \th))\\ 
\label{eq:adjoint3}
d_\th l(y, D(z^*, \th))= \del_\th l(y, D(z^*, \th))
- \int_0^{\t} \a(t) \lambda^{T} \del_\th  \nabla_z l(z(t),\th) dt, 
\end{align}
where we used $z^* = z(\t)$. 
Equations [\ref{eq:adjoint1}-\ref{eq:adjoint3}] define the so called adjoint method for gradient flow optimization of the loss. 
Due to the cost of solving all three equations, we empirically find that for this work sufficient and efficient optimization can be accomplished by ignoring the integral (``adjoint function'') part of the method. Theoretically, this is equivalent to ignoring the higher order term of the total differential $d_\th l(y, D(z^*, \th))= \del_\th l(y, D(z^*, \th)) + \del_z l(y, D(z^*, \th)) \del_\th z \approx \del_\th l(y, D(z^*, \th))$. Reducing Equations [\ref{eq:adjoint1}-\ref{eq:adjoint3}] to:
\begin{align}
\label{eq:approx1}
d z/d t = - \a(t) \nabla_z l(y, D(z(t), \th)), \text{ with } z(0)=0 \\
\label{eq:approx2}
d_\th \mathcal{L}= \del_\th l(y, D(z^*, \th)),
\end{align}
i.e. optimise the latent space via solving an ordinary differential equation and minimise the loss ``naively'' with respect to the parameters ignoring the dependence of $z(\t)$ to $\th$.

\begin{algorithm}[ht!]
   \caption{training GFE-amd for one sample per batch}
   \label{alg:example}
\begin{algorithmic}
   \State {\bfseries Input: } z(0)
   \State {\bfseries Output:} {$\mathbf{y}_b$, $\mathbf{\theta}_D$, $\mathbf{z}^*_b$} 
   \For{$b \in \{1,2,...,B\}$}{
   \While {$\frac{d}{db} {L}_{CE}(b) < 0.01$}{
   \While {$l(z({t}^n_{b} + \delta {t}^n_{b}) < l(z({t}^n_{b}$))}{
   \State $z({t}^{n+1}_{b}) \leftarrow z({t}^{n}_{b}) + [-\nabla_z l(z({t}^{n}_{b}))]\delta {t}^n_{b}$
   \State {$\delta {t}^{n+1}_{b} \leftarrow \beta^r\delta {t}^n_{b}$}
   \State {$r \leftarrow r+1$}
    }
    \State {$n \leftarrow n+1$}
   }
   \State {$z^*_b \leftarrow z({t}^n_{b})$}
   \State {$L(\theta_D) \leftarrow l(y, D(z^*, \theta_D)$}
   \State {$\theta_D \leftarrow$ backpropagate $L(\theta_D)$}
    }
\end{algorithmic}
\end{algorithm}

\subsection{Fixed grid ODE solver \label{app:fixed}}
Being an ODE, the gradient flow can be solved with general ODE solvers. Unfortunately, generic adaptive step size solvers are not useful because the underlying ODE becomes stiff quickly during training, the step size is reduced to extremely small values and the time it takes to solve gradient flow ODEs at each iteration takes an exorbitant amount of time. Fixed time-step or time grid solvers can be used, despite the stiffness. However, we empirically observed that these schemes can lead to instabilities in the training, see Fig.~\ref{fig:all_methods}. To demonstrate this, we experimented with a $4^{th}$ order Runge-Kutta method with fixed step size. The $\d t$ slices are predefined in logarithmic series such as $\d t$ is smaller closer to $t=0$, where integrands, $-\nabla_z l(y,D(z,\th))$, are more rapidly changing. Similarly, $\a$ is empirically set to $e^{(-2t/\t)}$ to facilitate faster convergence of $z$.
For the GFE full adjoint method, the integrands for each time slice are saved during the forward pass so they can be used for the calculation of the adjoin variable $\l$ in the backward pass. We used the same strategy for both basic gradient flow and the $2^{nd}$ order model.


\section{Experiments on the proposed methods \label{app:experiments}}

For training with MNIST and FashionMNIST datasets, we implement a sequential linear neural network. The decoder network architecture corresponds to four gradually increasing linear layers, with ELU non-linearities in-between. The exact reverse is used for the encoder of the AE. 


The network training is carried out with a momentum gradient decent optimiser (RMSprop), learning rate $0.0005$, $\e=1 \times 10^{-6}$ and $\a=0.9$. The GFE and AE are considered trained after one epoch and twelve epochs respectively. 

\begin{figure}[ht]
\begin{center}
\includegraphics[width=1\columnwidth]{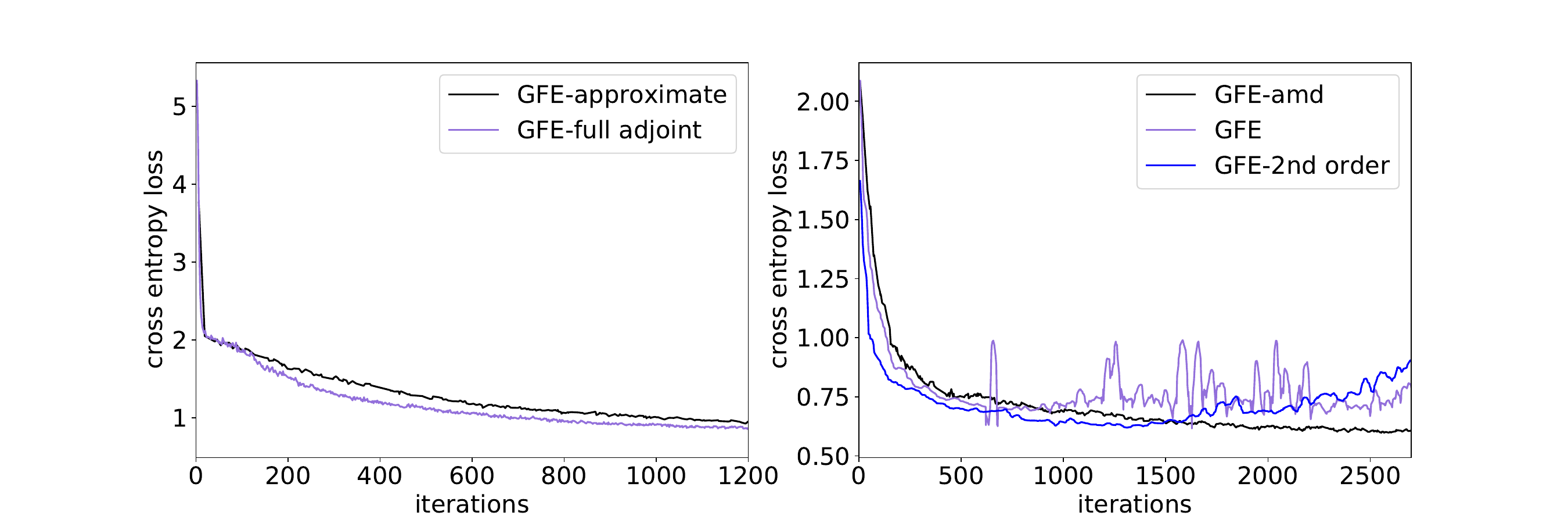}
\caption{\label{fig:all_methods} \textbf{Left} Validation mean cross-entropy loss plotted against MNIST training iterations for the approximate and full adjoint GFE methods. The full adjoint has a slight advantage over the approximate. \textbf{Right} Validation mean cross-entropy loss plotted against MNIST training iterations for the GFE, $2^{nd}$ order GFE and GFE-amd methods. The GFE-amd is both more stable and approaches a better convergence relative to the other methods}
\end{center}
\end{figure}

Initially a relative comparison between the full adjoint and the approximate fixed grid GFE methods is carried out to assess the relevance of the higher order term. Specifically, we carry out MNIST experiments for a fixed network random seed, where we  trained the Decoder using the different GFE methods and computed cross entropy loss over the validation set. The proper adjoint solution requires Equations \ref{eq:adjoint1} and \ref{eq:adjoint2} to be solved for each slice of the integral in Eq. \ref{eq:adjoint3}. Given $N$ time-slices (for sufficient accuracy $N\approx 100$), this requires $\mathcal{O}(5N)$ calls to the model $D$ for each training image.  The approximate method as in Equations \ref{eq:approx1} and \ref{eq:approx2} requires only $\mathcal{O}(N)$ passes. From Fig. \ref{fig:all_methods}~(left) it is evident that the 5-fold increase in computational time is not cost-effective as the relative reduction in loss convergence per iteration is not significant. 

Furthermore, to increase convergence with respect to training data, the accelerated gradient flow $2^{nd}$ order GFE is implemented in Section \ref{sec:nesterov}.
From Fig. \ref{fig:all_methods}~(right),  the accelerated gradient method increases initially the convergence per iteration relative to GFE, nevertheless it is slightly more computationally expensive due to solving a coupled system. Additionally, from the same Fig. certain stability issues are observed for both GFE and second order GFE methods later on despite the initial efficient learning. In order to guarantee  stability, the GFE-amd method is implemented as explained in Section \ref{sec:amd}. The black curve in Fig. \ref{fig:all_methods}~(right) shows a clear improvement of the GFE-amd over the later methods. Importantly, this result is robust to O of the experiment.

\section{Further results}
\begin{figure}[ht]
\begin{center}
\includegraphics[width=\columnwidth]{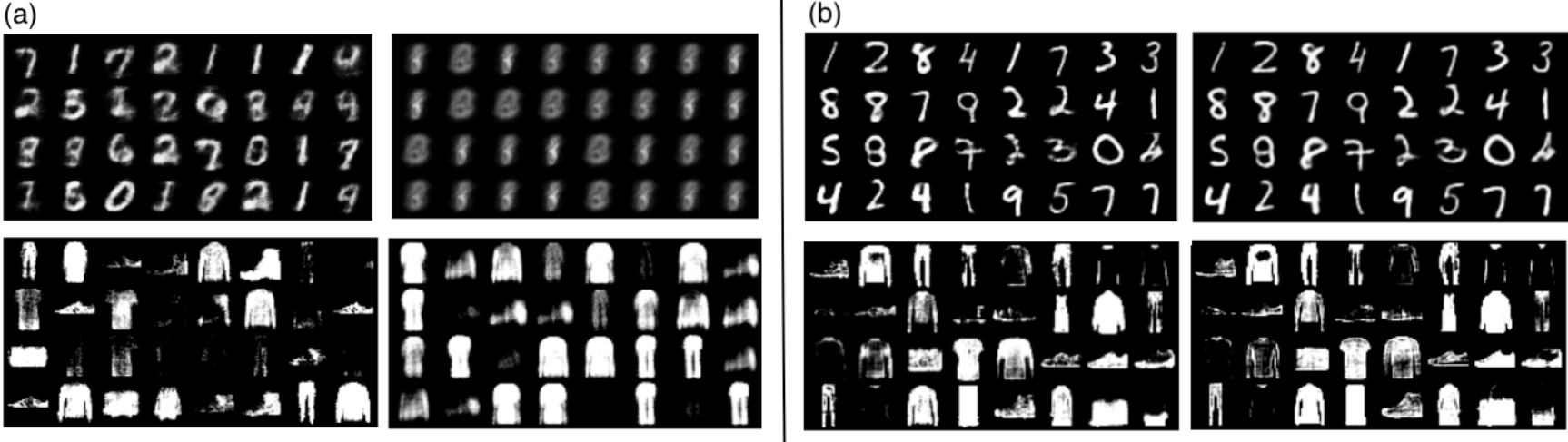}
\end{center}
\caption{\label{fig:images} \textbf{(a)} Test-set reconstructions for trained GFE-amd (left) and AE (right)  that only see  $1\%$ of MNIST (top) and FashionMNIST (bottom) training images. \textbf{(b)} Test-set reconstructions for fully trained GFE-amd (left) and AE (right)  with  MNIST (top) and FashionMNIST (bottom) training images. Note: The labels are identical in the respective reconstructions.}
\end{figure}

Sample test-set reconstructions with a fixed network random seed for GFE-amd and AE methods are shown in Fig. \ref{fig:images}. From Fig. \ref{fig:images}~(a) it is evident that the GFE-amd is superior in producing accurate reconstructions with the very limited amount of data. Fig. \ref{fig:images}~(b) indicates that both GFE-amd and AE generate similar reconstructions when properly trained. 

\end{document}

%% file: pakete.tex
\usepackage[T1]{fontenc}

\usepackage[ngerman,british]{babel}
	\addto\captionsngerman{}

\usepackage{amsmath,amssymb,amsthm,dsfont,mathtools,mathrsfs,bbm}

\usepackage{mdwlist}
\usepackage{paralist}
\usepackage{array}
\usepackage{booktabs}						
\usepackage{dcolumn} 						

\usepackage{graphicx}
\usepackage{wrapfig}

\usepackage{url}

\usepackage{color}
	\definecolor{myblue}{rgb}{0,0.3,0.8}
	\definecolor{mygreen}{rgb}{0,0.5,0}
	\definecolor{myblue}{rgb}{0,0.3,0.8}

\usepackage[version=3]{mhchem}

\usepackage{savesym}



%% file: befehle.tex
\savesymbol{a}
\savesymbol{b}
\savesymbol{d}
\savesymbol{e}
\savesymbol{k}
\savesymbol{l}
\savesymbol{H}
\savesymbol{L}
\savesymbol{p}
\savesymbol{r}
\savesymbol{s}
\savesymbol{S}
\savesymbol{F}
\savesymbol{t}
\savesymbol{w}
\savesymbol{th}
\savesymbol{div}
\savesymbol{tr}
\savesymbol{Re}
\savesymbol{Im}
\savesymbol{AA}
\savesymbol{SS}
\savesymbol{Si}
\savesymbol{gg}
\savesymbol{ss}

\newcommand{\a}{\alpha}

\newcommand{\e}{\varepsilon}
\newcommand{\d}{\delta}
\newcommand{\del}{\partial}

\newcommand{\l}{\lambda}

\newcommand{\t}{\tau}
\newcommand{\th}{\theta}



%% file: makros.tex
\newlength{\tmplen}

